\documentclass[accepted]{uai2023} % after acceptance, for a revised
\usepackage[american]{babel}

\usepackage{natbib} % has a nice set of citation styles and commands
\usepackage{mathtools} % amsmath with fixes and additions
\usepackage{booktabs} % commands to create good-looking tables
\usepackage{algorithm}
\usepackage{algorithmic}
\usepackage{amsmath}
\usepackage{amsthm}
\usepackage{amssymb}
\theoremstyle{definition}
\newtheorem{definition}{Definition}[section]
\theoremstyle{plain}
\newtheorem{theorem}{Theorem}[section]

\usepackage{tikz}
\usetikzlibrary{arrows,automata}
\usepackage[capitalize,noabbrev]{cleveref}
\usepackage{nicefrac}
\usepackage{booktabs}

\hypersetup{colorlinks=true,citecolor=darkgray,linkcolor=black,urlcolor=black}

\theoremstyle{plain}

\newtheorem{corollary}[theorem]{Corollary}
\theoremstyle{remark}

\crefname{equation}{Eq.}{Eqs.}

\newcommand{\piwsu}{\hat{\pi}}
\newcommand{\Real}{\mathbb{R}}
\newcommand{\opt}{^{*}}
\renewcommand{\P}{\mathbb{P}}
\newcommand{\PiR}{\Pi_{\mathrm{R}}}
\newcommand{\PiH}{\Pi_{\mathrm{H}}}

\DeclareMathOperator*{\argmax}{argmax}

\renewcommand{\cite}[1]{\citep{#1}}

\crefname{equation}{}{} 
\usepackage{titlesec}
\titlespacing*{\paragraph} {0pt}{0.1ex plus 0.1ex minus .05ex}{0.5em}

\title{Solving Multi-Model MDPs by Coordinate Ascent and Dynamic Programming}

\author[1]{\href{mailto:<xihong.su@unh.edu>?Subject=UAI2023}{Xihong Su}{}}
\author[1]{Marek Petrik}
\affil[1]{%
    Department of Computer Science \\
    University of New Hampshire \\
    Durham, NH, USA
}
\begin{document}
\maketitle

\begin{abstract}
Multi-model Markov decision process~(MMDP) is a promising framework for computing policies that are robust to parameter uncertainty in MDPs. MMDPs aim to find a policy that maximizes the expected return over a \emph{distribution} of MDP models. Because MMDPs are NP-hard to solve, most methods resort to approximations. In this paper, we derive the policy gradient of MMDPs and propose CADP, which combines a coordinate ascent method and a dynamic programming algorithm for solving MMDPs. The main innovation of CADP compared with earlier algorithms is to take the coordinate ascent perspective to adjust model weights iteratively to guarantee monotone policy improvements to a local maximum. A theoretical analysis of CADP proves that it never performs worse than previous dynamic programming algorithms like WSU. Our numerical results indicate that CADP substantially outperforms existing methods on several benchmark problems.
\end{abstract}

\section{Introduction}

Markov Decision Processes~(MDPs) are commonly used to model sequential decision-making in uncertain environments, including reinforcement learning, inventory control, finance, healthcare, and medicine~\cite{puterman2014markov,boucherie2017markov,Sutton2018}. In most applications, like reinforcement learning~(RL), parameters of an MDP must be usually estimated from observational data, which inevitably leads to model errors. Model errors pose a significant challenge in many practical applications. Even small errors can accumulate, and policies that perform well in the estimated model can fail catastrophically when deployed~\cite{steimle2021multi,behzadian2021optimizing,petrik2019beyond,nilim2005robust}. Therefore, it is important to develop algorithms that can compute policies that are reliable even when the MDP parameters, such as transition probabilities and rewards, are not known exactly.

Our goal in this work is to solve finite-horizon \emph{multi-model MDPs}~(MMDPs), which were recently proposed as a viable model for computing reliable policies for sequential decision-making problems~\cite{buchholz2019computation,steimle2021multi,ahluwalia2021policy,Hallak2015a}. MMDPs assume that the exact model, including transition probabilities and rewards, is unknown, and instead, one possesses a \emph{distribution} over MDP models. Given the model distribution, the objective is to compute a \emph{Markov} (history-independent) policy that maximizes the return averaged over the uncertain models. MMDPs arise naturally in multiple contexts because they can be used to minimize the expected Bayes regret in \emph{offline reinforcement learning}~\cite{steimle2021multi}.

Because solving MMDPs optimally is NP-hard~\cite{steimle2021multi,buchholz2019computation}, most algorithms compute approximately-optimal policies. One line of work has formulated the MMDP objective as a \emph{mixed integer linear program}~(MILP)~\cite{buchholz2019computation,lobo2020soft,steimle2021multi}. MILP formulations can solve small MMDPs optimally when given sufficient time, but they are hard to scale to large problems~\cite{ahluwalia2021policy}. Another line of work has sought to develop \emph{dynamic programing} algorithms for MMDPs~\cite{steimle2021multi,lobo2020soft,buchholz2019computation,steimle2018multi}. Dynamic programming formulations lack optimality guarantees, but they exhibit good empirical behaviors and can be used as the basis for scalable MMDP algorithms that leverage reinforcement learning or value functions, or policy approximations.

In this paper, we identify a new connection between policy gradient and dynamic programming in MMDPs and use it to introduce \emph{Coordinate Ascent Dynamic Programming}~(CADP) algorithm. CADP improves over both dynamic programming and policy gradient algorithms for MMDPs. In fact, CADP closely resembles the prior state-of-the-art dynamic programming algorithm, Weight-Select-Update~(WSU)~\cite{steimle2021multi}, but uses adjustable model weights to improve its theoretical properties and empirical performance. Compared with generic policy gradient MMDP algorithms, CADP reduces the computational complexity, provides better theoretical guarantees and better empirical performance. 

Although we focus on tabular MMDPs  in this work, algorithms that combine dynamic programming with policy gradient, akin to actor-critic algorithms, have an impressive track record in solving large and complex MDPs in reinforcement learning. Similarly to actor-critic algorithms, the ideas that underlie CADP generalize readily to large problems, but the empirical and theoretical analysis of such approaches is beyond the scope of this paper. It is also important to note that one cannot expect the policy gradient in MMDP to have the same properties as in ordinary MDPs. For example, recent work shows that policy gradient converges to the optimal policy in tabular MDPs~\cite{bhandari2021linear,agarwal2021theory}, but one cannot expect the same behavior in tabular MMDPs because this objective is NP-hard.

Finally, our CADP algorithm can serve as a foundation for new robust reinforcement learning algorithms. Most popular robust reinforcement learning algorithms rely on robust MDPs in some capacity. Robust MDPs maximize returns for the \emph{worst} plausible model error~(e.g.,~\cite{iyengar2005robust,ho2021a,goyal2022robust}) but are known to generally compute policies that are overly conservative. This is a widely recognized problem, and several recent frameworks attempt to mitigate it, such as \emph{percentile-criterion}~\cite{Delage2009,behzadian2021optimizing}, \emph{light-robustness}~\cite{Buchholz2019}, \emph{soft-robustness}~\cite{Derman2018,lobo2020soft}, and \emph{distributional robustness}~\cite{Xu2012}. \emph{Multi-model MDPs} can be seen as a special case of light-robustness and soft-robustness, and CADP can be used, as we show below, to improve some of the existing algorithms proposed for these general objectives.  

The remainder of the paper is organized as follows. We discuss related work in \cref{sec:related-work}. The multi-model MDP framework is defined in \cref{sec:framework}. In \cref{sec:solution-methods}, we derive the policy gradient of MMDPs and present the CADP algorithm, which we then analyze theoretically in \cref{sec: theoretical-analysis}. Finally, \cref{sec:numer-exper} evaluates CADP empirically. % and shows that it outperforms previous dynamic programming algorithms, policy gradient algorithms, and other relevant approaches. 

\section{Related Work} \label{sec:related-work}

Numerous research areas have considered formulations or goals closely related to the MMDP model and objectives. In this section, we briefly review the relationship of MMDPs with other models and objectives; given the breadth and scope of these connections, it is inevitable that we omit some notable but only tangentially relevant work.

\paragraph{Robust and soft-robust MDPs}

Robust optimization is an optimization methodology that aims to reduce solution sensitivity to model errors by optimizing for the worst-case model error~\cite{Ben-Tal2009}. Robust MDPs use the robust optimization principle to compute policies to MDPs with uncertain transition probabilities~\cite{nilim2005robust,iyengar2005robust,wiesemann2013robust}. However, the max-min approach to model uncertainty MDPs has been proposed several times before under various names~\cite{Satia1973,Givan2000}. Robust MDPs are tractable under an independence assumption popularly known as \emph{rectangularity}~\cite{iyengar2005robust,wiesemann2013robust,petrik2019beyond,goyal2022robust,mannor2016robust}. Unfortunately, rectangular MDP formulations tend to give rise to overly conservative policies that achieve poor returns when model errors are small. \emph{Soft-robust}, \emph{light-robust}, and \emph{distributionally-robust} objectives assume some underlying Bayesian probability distribution over uncertain models and use risk measures to balance the average and the worst returns better~\cite{Xu2012,lobo2020soft,Derman2018,Delage2009,Satia1973}. Unfortunately, virtually all of these formulations give rise to intractable optimization problems. 

\paragraph{Multi-model MDPs}

MMDP is a recent model that can be cast as a special case of soft-robust optimization~\cite{steimle2021multi,buchholz2019computation}. The MMDP formulation also assumes a Bayesian distribution over models and seeks to compute a policy that maximizes the average return across these uncertain models~\cite{buchholz2019computation}. Even though optimal policies in these models may need to be history-dependent, the goal is to compute \emph{Markov policies}. Markov policies depend only on the current state and time step and can be more practical because they are easier to understand, analyze, and implement~\cite{Petrik2016}. Existing MMDP algorithms either formulate and solve the problem as a mixed integer linear program~\cite{buchholz2019computation,steimle2021decomposition,ahluwalia2021policy}, or solve it approximately using a dynamic programming method, like WSU~\cite{steimle2021multi}.

\paragraph{POMDPs}

One can formulate an MMDP model as a \emph{partially observable MDP}~(POMDP) in which the hidden states represent the uncertain model-state pairs ~\cite{kaelbling1998planning,steimle2021multi,buchholz2019computation}. Most POMDP algorithms compute history-dependent policies~\cite{kochenderfer2022algorithms}, and therefore, are not suitable for computing Markov policies for MMDPs~\cite{steimle2021multi}. On the other hand, algorithms for computing finite-state controllers in POMDPs~\cite{Vlassis2012} or implementable policies~\cite{Petrik2016,ferrer2020solving} compute stationary policies. Stationary policies are inappropriate for the finite-horizon objectives that we target.

\paragraph{Bayesian Multi-armed Bandits}

MMDPs are also related to Bayesian exploration and multi-armed bandits. Similarly to MMDPs, \emph{Bayesian exploration} seeks to minimize the Bayesian regret, which is computed as the average regret over the unknown MDP model~\cite{lattimore2020bandit}. Most research in this area has focused on the bandit setting, which corresponds to an MDP with a single state. \emph{MixTS} is a recent algorithm that generalizes Thompson sampling to the full MDP case~\cite{hong2022thompson}. MixTS achieves a sublinear regret bound but computes policies that are history dependent and, therefore, are not Markov. We include MixTS in our empirical comparison and show that Markov policies cannot achieve sublinear Bayes regret. 

\paragraph{Policy Gradient}

As discussed in the introduction, CADP combines policy gradient methods with dynamic programming. Policy gradient methods are widespread in reinforcement learning and take a first-order optimization to policy improvement. Many policy gradient methods are known to be adaptations of general unconstrained or constrained first-order optimization algorithms---such as Frank-Wolfe, projected gradient descent, mirror descent, and natural gradient descent---to the return maximization problem~\cite{bhandari2021linear}. CADP builds on these methods but uses dynamic programming to perform more efficient gradient updates reusing much more prior information than the generic techniques.

\section{Framework: Multi-model MDPs}\label{sec:framework}

In this section, we formally describe the MMDP framework and show how it arises naturally in Bayesian regret minimization. We also summarize WSU, a state-of-the-art dynamic programming algorithm, to illustrate the connections between CADP and prior dynamic programming algorithms. 

\paragraph{MMDPs}
A \emph{finite-horizon MMDP} comprises the horizon $\mathcal{T}$, states $\mathcal{S}$, actions $\mathcal{A}$, models $\mathcal{M}$, transition function $p$, rewards $r$, initial distribution $\mu$, model distribution $\lambda$~\cite{steimle2021multi}. The symbol $\mathcal{T} = \{1, \dots , T\}$ is the set of decision epochs, $\mathcal{S} = \{1, \dots , S\}$ is the set of states, $\mathcal{A} = \{ 1, \dots , A \}$ is the set of actions, and $\mathcal{M} = \{ 1, \dots , M \}$ is the set of possible  models. The function $p^m\colon \mathcal{S} \times \mathcal{A} \to \Delta^{\mathcal{S}},\, m \in \mathcal{M}$ is the transition probability function, which assigns a distribution from the $S$-dimensional probability simplex $\Delta^{S}$ to each combination of a state, an action, and a model $m$ from the finite set of models $\mathcal{M}$. The functions $r^m_t\colon \mathcal{S} \times \mathcal{A} \to \Real, m\in \mathcal{M}, t\in  \mathcal{T}$ represent the reward functions. $ \mu $ is the initial distribution over states. Finally, $\lambda = (\lambda_{1}, \dots, \lambda_{M} )$ represents the set of initial model probabilities (or weights) with $\lambda_{m} \in (0,1)$ and $ \sum_{m \in \mathcal{M}} \lambda_{m} = 1$.

Note that the definition of an MMDP does not include a discount factor. However, one could easily adapt the framework to incorporate the discount factor $\gamma \in [0,1]$. It is sufficient to define a new time-dependent reward function $\hat{r}^{m}_{t} = \gamma^{t-1} r^{m}_{t}$ and solve the MMDP with this new reward function. 

Before describing the basic concepts necessary to solve MMDPs, we briefly discuss how one may construct the models $\mathcal{M}$ and their weights $\lambda_m$ in a practical application. The model weights $\lambda_{m} \in (0,1)$  may be determined by expert judgment, estimated from empirical distributions using Bayesian inference, or treated as uniform priors~\cite{steimle2021multi}. Prior work assumes that the decision maker has accurate estimates of model weights $\lambda_{m}$ or treats model weights as uniform priors~\cite{steimle2021multi,bertsimas2018optimal}. Once $\lambda_{m}, m \in \mathcal{M}$ is specified, the value of $\lambda_{m}$ does not change.

The solution to an MMDP is a \emph{deterministic Markov policy} $\pi_t \colon \mathcal{S} \to  \mathcal{A} ,\, t\in \mathcal{T}$ from the set of deterministic Markov policies $\Pi$. A policy $\pi_t(s)$ prescribes which action to take in time step $t$ and state $s$. It is important to note that the action is non-stationary---it depends on $t$---but it is independent of the model and history. The policies for MMDPs, therefore, mirror the optimal policies in standard finite-horizon MDPs. To derive the policy gradient, we will also need randomized policies $\PiR$ defined as $\pi_t\colon \mathcal{S} \to \Delta^{\mathcal{A}}, t\in \mathcal{T}$.

The \emph{return} $\rho\colon \Pi \to \Real$ for each policy $\pi\in \Pi$ is defined as the \emph{mean} return  across the uncertain true models:\begin{equation}\label{eq:return}
  \rho(\pi) =
  % \sum_{m \in\mathcal{M}} \lambda_{m} \cdot \mathbb{E}^{\pi,p^{m},\mu} \left[ \sum_{t=1}^{T}  r_{t}^{m}(S_{t},A_{t})\right] ~.
\mathbb{E}^{\lambda} \left[  \mathbb{E}^{\pi,p^{\tilde{m}},\mu} \left[ \sum_{t=1}^{T}  r_{t}^{\tilde{m}}(\tilde{s}_{t},\tilde{a}_{t}) \mid \tilde{m} \right] \right]~.
\end{equation}
$\tilde{m}, \tilde{s_t}$ and $\tilde{a_t}$ are random variables. 

The decision maker seeks a policy that maximizes the return:
\begin{equation}\label{eq:objective-return}
  \rho\opt \; =\;  \max_{\pi \in \Pi} \rho(\pi)~.
\end{equation}
As with prior work on MMDPs, we restrict our attention to deterministic Markov policies because they are easier to compute,
analyze, and deploy. While history-dependent policies can, in principle, achieve better returns than Markov policies, our numerical results show that existing state-of-the-art algorithms compute history-dependent policies that are inferior to the Markov policies computed by CADP.

Next, we introduce some quantities that are needed to describe our algorithm. Because an MMDP for a fixed model $m \in \mathcal{M}$ is in ordinary MDP, we can define the value function $v_{t,m}^{\pi}\colon \mathcal{S}\to \Real$ for each $\pi\in \Pi$, $t\in \mathcal{T}$, and $s\in \mathcal{S}$ as \cite{puterman2014markov}
\begin{equation}\label{eq:v-t-m-s}
v_{t,m}^{\pi}(s) = \mathbb{E}\left[\sum_{t' =t}^{T}r_{t'}^m(\tilde{s}_{t'},\tilde{a}_{t'}) \mid \tilde{s}_t = s, \tilde{m} = m\right].
\end{equation}
The value function also satisfies the \emph{Bellman equation} 
\begin{equation}\label{eq:backup}
v_{t,m}^{\pi}(s) = \sum_{a\in \mathcal{A}}  \pi_t(s,a)\cdot q_{t,m}^{\pi}(s,a) \,,
\end{equation}  
where state-action value function $q^{\pi}_{t,m}\colon \mathcal{S}\times \mathcal{A} \to \Real$ is defined as
\begin{equation}\label{eq:value-0f-(state-action)}
q_{t,m}^\pi (s, a) = r_t^{m}(s,a)  + \sum_{s' \in \mathcal{S}} p_t^m(s' \mid  s,a) \cdot v_{t+1,m}^{\pi}(s').
\end{equation}
The optimal value function $v\opt_{t,m}\colon \mathcal{S} \to \Real$ is the value function of the optimal policy $\pi\opt_m \in \Pi$ and satisfies that
\[
 v\opt_{t,m}(s) = \max_{a\in \mathcal{A}} q\opt_{t,m}(s,a), \quad  s\in \mathcal{S},
\]
where $q\opt_{t,m} = q_{t,m}^{\pi\opt_m}$.

Unlike in an MDP, the value function does \emph{not} represent the value of being in a state because the model $m$ is unknown.

\paragraph{Bayesian regret minimization}
The multiple models in MMDPs can originate from various sources~\cite{steimle2021multi}. We now briefly describe how these models can arise in offline RL because this is the setting we focus on in the experimental evaluation. In offline RL, the decision maker needs to compute a policy $\pi \in \Pi$ using a logged dataset of state transitions $\mathcal{D} = (t, s_i, a_i, s_i')$. In Bayesian offline RL, the decision maker is equipped with a prior distribution $\kappa \in \Delta^{\mathcal{M}}$ over the (possibly infinite) set of models $\mathcal{M}$ and uses the data $\mathcal{D}$ to compute a posterior distribution $\lambda \in \Delta^{\mathcal{M}}$. The goal is then to find a policy $\hat{\pi}\in \Pi$ that minimizes the Bayes regret
\begin{equation}\label{eq:regret}
 \hat{\pi} \in \arg \min_{\pi \in \Pi}  
\mathbb{E}^{\lambda} \left[\rho^{\tilde{m}}(\pi\opt_{\tilde{m}}) - \rho^{\tilde{m}}(\pi) \right]\,,
\end{equation}
where $\rho^m\colon \Pi \to \Real$ for $m\in \mathcal{M}$ is the return for the model $m$ and each policy. Note that $\tilde{m}$ is the random variable that represents the model. A policy is optimal in ~\eqref{eq:regret} if and only if it is optimal in \eqref{eq:objective-return} because the expectation operator is linear. The minimum regret policy can then be approximated by an MMDP using a finite approximation of the posterior distribution $\lambda$.

\paragraph{Dynamic Programming Algorithms}

The simplest dynamic program algorithm for an MMDP is known as Mean Value Problem~(MVP)~\cite{steimle2021multi}. MVP first computes an average transition probability function $\bar{p}_t\colon  \mathcal{S} \times \mathcal{A} \to  \Delta^{\mathcal{S}}, t\in \mathcal{T}$ for each $s,s'\in \mathcal{S}$, $a\in \mathcal{A}$ as
\[
 \bar{p}_t(s' \mid s,a) = \sum_{m\in \mathcal{M}} \lambda_m \cdot p_t^m(s'\mid s,a)~.
\]
and the average reward function $\bar{r}_t \colon \mathcal{S} \times \mathcal{A} \to \Real, t\in \mathcal{T}$ as
\[
 \bar{r}_t( s,a) = \sum_{m\in \mathcal{M}} \lambda_m\cdot r_t^m(s,a)~.
\]
One can then compute a policy $\bar{\pi}\in \Pi$ by solving the MDP with the transition function $\bar{p}$ and a reward function $\bar{r}$ using standard algorithms~\cite{puterman2014markov}.

A more sophisticated dynamic programming algorithm, WSU, significantly improves over MVP~\cite{steimle2021multi}. WSU resembles value iteration and updates the policy $\piwsu_t\colon \mathcal{S} \to \mathcal{A}$ and state-action value function $v^{\piwsu}_{t,m}$ for each model $m\in \mathcal{M}$ backward in time. After initializing the value function $v_{t,m}^{\piwsu}$ to $0$ at time $t = T$ for each $m\in \mathcal{M}$, it computes $q_{t,m}^{\piwsu}$ from \cref{eq:value-0f-(state-action)} at time $t = T-1$. The policy $\piwsu_t$ at time $t = T-1$ is computed by solving the following optimization problem:
\begin{equation}\label{eq:wsu-optimal-action}
  \piwsu_t(s_t) \in \arg\max_{a \in \mathcal{A}} \sum_{m \in \mathcal{M}} \lambda_m \cdot q_{t,m}^{\piwsu}(s_t, a),  \quad \forall s_t\in \mathcal{S}.
\end{equation}
The optimization in \cref{eq:wsu-optimal-action} chooses an action that maximizes the weighted sum of the values of the individual models~\cite{steimle2021multi}. After computing $\piwsu_t$ for $t=T-1$, WSU repeats the procedure for $T-2, T-3, \dots, 1$.

It is essential to discuss the limitations of WSU that CADP improves on. At each time step $t\in T, T-1, \dots , 1$, WSU computes the policy $\piwsu_t$ that maximizes the sum of values of models weighted by the \emph{initial} weights $\lambda_1, \dots , \lambda_M$. But using the initial weights $\lambda_1, \dots , \lambda_M$ here is not necessarily the correct choice. Recall that each model $m$ has potentially different transition probabilities. Simply being a state $s_t$ reveals some information about which models are more likely. One should not use the prior distribution $\lambda_1, \dots , \lambda_M$, but instead the \emph{posterior} distribution conditional on being in state $s_t$. This is what CADP does, and we describe it in the next section. 

\section{CADP Algorithm}
\label{sec:solution-methods}

We now describe CADP, our new algorithm that combines coordinate ascent with dynamic programming to solve MMDPs. CADP differs from WSU in that it appropriately adjusts model weights in the dynamic program.

In the remainder of the section, we first describe adjustable model weights in \cref{sec:model-weights}. These weights are needed in deriving the MMDP policy gradient in \cref{sec:policy-grad-mmdps}. Finally, we describe the CADP algorithm and its relationship to coordinate ascent in \cref{sec:algorithm}.

\subsection{Model Weights} \label{sec:model-weights}

We now give the formal definition of model weights. Informally, a model weight $b_{t,m}^{\pi}(s)$ represents the \emph{joint} probability of $m$ being the true model and the state at time $t$ being $s$ when the agent follows a policy $\pi$. The value $b_{t,m}^{\pi}(s)$ is useful in expressing the gradient of $\rho(\pi)$.

\begin{definition} \label{def-belief-state}
An \emph{adjustable weight} for each model $m\in \mathcal{M}$, policy $\pi\in \Pi$, time step $t\in \mathcal{T}$ , and state $s \in \mathcal{S}$ is
\begin{equation} \label{eq:belief-state}
  b_{t,m}^{\pi}(s) \; =\;  \P\left[\tilde{m} = m, \tilde{s}_t= s\right]\,,
\end{equation}
where $S_0 \sim \mu$, $\tilde{m} \sim \lambda$, and $\tilde{s}_1, \dots , \tilde{s}_T$ are distributed according to $p^{\tilde{m}}$ of policy $\pi$.
\end{definition}

Although the model weight $b_{t,m}^{\pi}(s)$ resembles the belief state in the POMDP formulation of MMDPs, it is different from it in several crucial ways. First, the model weight represents the \emph{joint} probability of a model and a state rather than a conditional probability of the model given a state. Recall that in a POMDP formulation of an MMDP, the latent states are $\mathcal{M} \times \mathcal{S}$, and the observations are $\mathcal{S}$. Therefore, the POMDP belief state is a distribution over $\mathcal{M} \times \mathcal{S}$. Second, model weights are \emph{Markov} while belief states are history-dependent. This is important because we can use the Markov property to compute model weights efficiently.

Computing the model weights $b$ directly from \cref{def-belief-state} is time-consuming. Instead, we propose a simple linear-time algorithm. At the initial time step $t=1$, we have that
\begin{equation}\label{eq:belief-initial}
  b_{1,m}^{\pi}(s) =   \lambda_{m} \cdot \mu(s), \quad \forall m \in \mathcal{M}, s \in \mathcal{S}, \pi \in \Pi.
\end{equation}
The weights for any $t' = t + 1$ for any $t = 1, \dots, T-1$ and any $s'\in \mathcal{S}$ can than be computed as
\begin{equation}\label{eq:belief-update}
  b_{t',m}^{\pi}(s') = \sum_{s_{t},a \in \mathcal{S} \times \mathcal{A}}
  p^{m}(s' |  s_{t}, a)  \pi_{t}(s_{t}, a)  b_{t,m}^{\pi}(s_{t})\,.
\end{equation}
Intuitively, the update in \cref{eq:belief-update} computes the marginal probability of each state at $t+1$ given the probabilities at time $t$. Note that this update can be performed for each model $m$ independently because the model does not change during the execution of the policy. 

Note that the adjustable model weights $b_{t,m}^{\pi}(s')$ are Markov because we only consider Markov policies $\Pi$. As discussed in the introduction, we do not consider history-dependent policies because they can be much more difficult to implement, deploy, analyze, and compute.

\subsection{MMDP Policy Gradient} \label{sec:policy-grad-mmdps}

Equipped with the definition of model weights, we are now ready to state the gradient of the return with respect to the set of randomized policies.
\begin{theorem}\label{thm:policy gradient}
The gradient of $\rho$ defined in \cref{eq:return} for each $t \in \mathcal{T}$, $\hat{s}\in \mathcal{S}$, $\hat{a} \in \mathcal{A}$, and $\pi \in \PiR$ satisfies that
\begin{equation}\label{eq: gradient-deter}
  \frac{\partial \rho(\pi)}{\partial\pi_t(\hat{s},\hat{a})}\; =\;
  \sum_{m \in \mathcal{M}} b_{t,m}^{\pi}(\hat{s})\cdot q_{t,m}^{\pi}(\hat{s},\hat{a})\,,
\end{equation}
where $q$ and $b$ are defined in \cref{eq:value-0f-(state-action)} and \cref{eq:belief-update} respectively. 
\end{theorem}
Please see the appendix for the proof of this theorem.

\subsection{Algorithm} \label{sec:algorithm}

To formalize the CADP algorithm, we take a coordinate ascent perspective to reformulate the objective function $\rho(\pi)$. In addition to establishing a connection between optimization and dynamic programming, this perspective is very useful in simplifying the theoretical analysis of CADP in \cref{sec: theoretical-analysis}.

The return function $\rho(\pi)$ can be seen as a multivariate function with the policy at each time step seen as a parameter:
 \[
  \rho(\pi) = \rho(\pi_1, \dots,\pi_t, \dots, \pi_T)
 \]
where $\pi_t = \displaystyle [\pi_t(s_1,a_1), \dots,\pi_t(s_S,a_A)]$ for each $t \in \mathcal{T}$ with $s_i \in \mathcal{S}$ and $ a_j \in \mathcal{A}$. 

The coordinate ascent (or descent) algorithm maximizes $\rho(\pi)$ by iteratively optimizing it along a subset of coordinates at a time~\cite{Bertsekas2016nonlinear}. The algorithm is useful when optimizing complex functions that simplify when a subset of the parameter is fixed. \Cref{thm:policy gradient} shows that the return $\rho$ function has exactly this property. In particular, while $\rho$ is non-linear and non-convex in general, the following result states that the function is linear for each specific subset of parameters. 
\begin{corollary} \label{cor:ret-linear}
For any policy $\bar{\pi} \in \Pi$ and $t\in \mathcal{T}$, the function $\pi_t \mapsto \rho(\bar{\pi}_1, \dots , \pi_t, \dots , \bar{\pi}_T)$ is \emph{linear}.  
\end{corollary}
\begin{proof}
The result follows immediately from~\eqref{eq: gradient-deter}, which shows that $\partial \rho /\partial\pi_t(s,a) = \sum_{m \in \mathcal{M}} b_{t,m}^{\pi}(s)\cdot q_{t,m}^{\pi}(s,a)$ which is constant in $\pi_t(s,a)$ for each $s\in \mathcal{S}$, $a\in \mathcal{A}$, and $t\in \mathcal{T}$. Therefore, we have that $\partial^2 \rho /\partial\pi_t(s,a)^2 = 0$ and the function $\pi_t \mapsto \rho(\bar{\pi}_1, \dots , \pi_t, \dots , \bar{\pi}_T)$ is linear by the multivariate Taylor's theorem.
\end{proof}

Ordinary coordinate ascent applied to $\rho(\pi)$ proceeds as follows. It starts with an initial policy $\pi^0 = (\pi_1^0, \dots, \pi_T^0)$. Then, at each iteration $n = 1, \dots $, it computes $\pi^{n}$ from $\pi^{n-1}$ by iteratively solving the following optimization problem for each $t\in \mathcal{T}$:
\begin{equation}\label{eq:coordinate-descent}
    \pi_t^n \in \argmax_{\hat{\pi}_t \in \Real^{\mathcal{S} \times\mathcal{A}}}\rho(\pi_1^{n-1}, \dots, \hat{\pi}_t, \dots, \pi_T^{n}) 
\end{equation}
From \cref{cor:ret-linear}, this is a linear optimization problem constrained to a simplex for each state individually. Therefore, using the standard optimality criteria over a simplex~(e.g,~Ex.~3.1.2 in \cite{Bertsekas2016nonlinear}) we have that the optimal solution in \eqref{eq:coordinate-descent} for each $s\in \mathcal{S}$ satisfies that
\begin{equation}\label{eq:swsu-optimal-action}
 \pi_t^n(s)  \in \argmax_{a \in \mathcal{A}} \sum_{m \in \mathcal{M}} b_{t,m}^{\pi^{n-1}}(s)  \cdot q_{t,m}^{\pi^n} (s,a) .
\end{equation}
$\pi^n$ can be solved by enumerating over the finite set of actions. This construction ensures that we get a sequence of policies $\pi^0, \pi^1,\pi^2, \dots$ with non-decreasing returns:$\rho(\pi^0) \leq \rho(\pi^1) \leq \rho(\pi^2) \leq \dots$. 

\begin{algorithm}[tb]
\caption{OptimizePolicy} 
\label{alg:optimize-policy}
\textbf{Input}: MMDPs,  Model weights $b^{\pi^{n-1}}$ \\
\textbf{Output}: $\pi^n = (\pi_1^n,\ldots,\pi_T^n)$
\begin{algorithmic}[1]
   \STATE Initialize $v_{T+1,m}^{\pi^n}(s_{T+1}) = 0, 
     \forall m \in \mathcal{M} $ \\
    \STATE Initialize $\pi^n \gets \pi^{n-1}$ \\
   \FOR{$t = T, T-1, \dots , 1$}
   %\COMMENT{Solve \eqref{eq:coordinate-descent} for $t$} \\
   \FOR{Every state $s_t \in \mathcal{S}$}
      \STATE Update $\pi^n_t(s_t)$ according to \cref{eq:swsu-optimal-action} with $b^{\pi^{n-1}}_t(s_t)$ \label{line:wsu-equal}
        \STATE Update $v_{t,m}^{\pi^n}(s_t)$ according to \cref{eq:backup} for each $m \in \mathcal{M}$  \\
    \ENDFOR  
   \ENDFOR
\STATE \textbf{return} $\pi^n$
\end{algorithmic}
\end{algorithm}

While the coordinate ascent scheme outlined above is simple and theoretically appealing, it is computationally inefficient. The computational inefficiency arises because computing the weights $b$ and value functions $q$ necessary in~\eqref{eq:swsu-optimal-action} requires one to update the entire dynamic program. The coordinate ascent algorithms must perform this time-consuming update in every iteration. To mitigate this computational issue, CADP interleaves the dynamic program with the coordinate ascent steps so that we reduce the updates of $b$ and $v$ to a minimum.

Conceptually, CADP is composed of two main components. \Cref{alg:optimize-policy} is the inner component that uses dynamic programming to compute a policy ${\pi}^{n}$ for some given adjustable model weights $b^{\pi^{n-1}}$. This algorithm uses the value function of visiting a state from \cref{eq:swsu-optimal-action} to choose an action in each state that maximizes the expected value of the action for each $s_t\in \mathcal{S}$.

At any time step $t$, the maximization attempts to improve the action to take at time step $t$. A curious feature of the update in \cref{eq:swsu-optimal-action} is that it relies on two different policies, $\pi^{n-1}$ and $\pi^n$. This is because it assumes that the weights $b_{t'}$ are computed for $t' \le t$ using policy, $\pi^{n-1}$, which is the policy the decision maker follows up to time step $t$. The policy for $t > t'$ would have been updated using the dynamic program and, therefore, is denoted as $\pi^n$.

\begin{algorithm}[tb]
   \caption{CADP: Coordinate Ascent Dynamic Programming}
   \label{alg:swsu}
\textbf{Input}: MMDP, $\pi^0$\\
\textbf{Output}: Policy $\pi\in \Pi $
\begin{algorithmic}[1]
   \STATE $n \gets 0$ 
   \REPEAT
   \STATE $ n \gets n+1 $
   \STATE $ b^{\pi^{n-1}}\gets $ model weights from \cref{eq:belief-update} using $\pi^{n-1}$
   \STATE $\pi^n \gets $ OptimizePolicy (MMDP, $ b^{\pi^{n-1}}) $
   \UNTIL{$ \rho (\pi^n) = \rho (\pi^{n-1}) $}
\STATE \textbf{return} $\pi^n$
\end{algorithmic}
\end{algorithm}

The second component of CADP is described in \cref{alg:swsu}. This algorithm starts with some arbitrary policy $\pi^0$ and then alternates between computing the adjustable model weights and improving the policy using \cref{alg:optimize-policy}. The initial policy $\pi^0$ can be arbitrary and computed using an algorithm like MVP, WSU or a randomized policy.% We find that the initial policy has only a tiny effect on the ultimate quality of the computed policy.

A single iteration of CADP has the time complexity of $\mathcal{O}(T S^2 A M)$ similar to running value iteration for each one of the models. The number of iterations could be quite large. In the worst case, the algorithm may run an exponential number of iterations, visiting a significant fraction of all deterministic policies. We show, however, in the following section that the algorithm cannot loop and that each iteration either terminates or computes a better policy. In contrast, the complexity of a plain coordinate ascent iteration over all parameters would be $\mathcal{O}(T^2 S^3 A^2 M)$.

It is also interesting to contrast CADP with WSU. Recall that the limitation of WSU stems from the fact that \cref{eq:wsu-optimal-action} relies on the \emph{initial} model weights that do not depend on the current state and time. We propose to use instead \emph{adjustable} model weights, which represent the joint probability of the current state and the model at each time step $t$. The following sections show that using these adjustable weights enables CADP's favorable theoretical properties and improves empirical solution quality.

\section{Error Bounds}\label{sec: theoretical-analysis}
In this section, we analyze the theoretical properties of CADP. In particular, we show that CADP will never decrease the return of the current policy. As a result, CADP can never cycle and terminates in a finite time. We also contrast MMDPs with Bayesian multi-armed bandits and show that one cannot expect an algorithm that computes Markov policies to achieve sublinear regret. 

The following theorem shows that the overall return does not decrease when the local value function does not decrease. 
\begin{theorem}\label{thm:improvement}
Suppose that \cref{alg:swsu} generates a policy $\pi^n=(\pi_t^n)^T_{t=1}$ at an iteration $n$, then \( \rho(\pi^{n}) \geq \rho(\pi^{n-1}).
\)
\end{theorem}
Please see the appendix for the proof of this theorem.

\Cref{thm:improvement} implies that \cref{alg:swsu} must terminate in finite time. This is because the algorithm either terminates or generates policies with monotonically increasing returns. With a finite number of policies, the algorithm must eventually terminate.
\begin{corollary}
\cref{alg:swsu} terminates in a finite number of iterations. 
\end{corollary}

Given that the iterations of CADP only improve the return of the policy, one may ask why the algorithm may fail to find the optimal policy. The reason is that the algorithm makes local improvements to each state. Finding the globally optimal solution may require changing the policy in two or more states simultaneously. The policy update in each one of the states may not improve the return, but the simultaneous update does. This property is different from the situation in MDPs, where the best action at time $t$ is independent of the actions at times $t' < t$. 

It is also important to acknowledge the limitations of our analysis. One could ensure that iterations of CADP do not decrease the policy's return by accepting improving policy changes only. CADP does better than this. It finds the improving changes and converges to a type of local maximum. The computed policy is a local maximum in the sense that no single-state updates can improve its return. 

Given the connection between MMDPs and Bayesian bandits, one may ask whether it is possible to give regret bounds on the policy computed by CADP. The main difference between CADP and multi-armed bandit literature is that we seek to compute Markov policies, while algorithms like Thompson sampling compute history-dependent policies. We show next that it is impossible to achieve guaranteed sublinear regret with Markov policies.

The regret of a policy $\pi$ is defined as the average performance loss with respect to the best possible policy:
\[
  R_T(\pi) = \max_{\bar\pi \in \PiH}\rho_T(\bar\pi) - \rho_T(\pi),
\]
where $\PiH$ is the set of all history-dependent randomized policies and $\rho_T$ is the return for the horizon of length $T$. 

The following theorem shows that it is impossible to achieve sublinear regret with Markov policies. 
\begin{theorem}\label{thm:no-sublinear-regret}
There exists an MMDP for which no Markov policy achieves sub-linear regret. That is, there exists no $\pi\in \Pi$, $c>0$, and $t'>0$ such that
\[
R_t(\pi) \le c\cdot t \quad \text{for all} \quad  t \ge t'\,.
\]
\end{theorem}
Please see the appendix for the proof of this theorem.

\section{Numerical Experiments}\label{sec:numer-exper}

\paragraph{Algorithms}In this section, we compare the expected return and runtime of CADP to several other algorithms designed for MMDPs as well as baseline policy gradient algorithms. We also compare CADP to related algorithms proposed for solving Bayesian multi-armed bandits and methods that reformulate MMDPs as POMDPs.

Our evaluation scenario is motivated by the application of MMDPs in Bayesian offline RL as described in \cref{sec:framework}. That is, we compute a posterior distribution over possible models $m\in \mathcal{M}$ using a dataset and a prior distribution. Then, we construct the MMDP by sampling \emph{training} MDP models from the posterior distribution. We evaluate the computed policy using a separate \emph{test} sample from the same posterior distribution. 

\paragraph{Domains}
\emph{Riverswim (RS)}: This is a larger variation~\cite{behzadian2021optimizing} of an eponymous domain proposed to test exploration in MDPs~\cite{strehl2008analysis}. The MMDP consists of 20 states, 2 actions, 100 training models, and 700 test models. The training models are used to compute the policy, and test models are used to evaluate its performance. As in machine learning, this helps to control over-fitting. The discount factor is 0.9. 

\emph{Population (POP)}: The population domain was proposed for evaluating robust MDP algorithms~\cite{petrik2019beyond}. It represents a pest control problem inspired by the types of problems found in agriculture. The MMDP consists of 51 states, 5 actions, 1000 training models, and 1000 test models. The discount factor is 0.9. \emph{Population-small (POPS)} is a variation of the same domain that comprises a limited set of 100 training models and 100 test models. 

\emph{HIV}: Variations of the HIV management domains have been widely used in RL literature and proposed to evaluate MMDP algorithms~\cite{steimle2021multi}. The parameter values are adapted from Chen et al.~\cite{chen2017sensitivity}, and the rewards are based on Bala et al.~\cite{bala2006optimal}. In this case study, the objective is to find the sequence that maximizes the expected total net monetary benefit~(NMB). The MMDP consists of 4 states, 3 actions, 50 training models, and 50 test models. The discount factor is 0.9.

\emph{Inventory (INV)}: This model represents a basic inventory management model in which the decision makers must optimize stocking models at each time step~\cite{ho2021a}. The MMDP consists of 20 states, 11 actions, 100 training models, and 200 test models. The discount factor is 0.95.

Our CADP implementation initializes the policy $\pi^0$ to the WSU solution, sets the weights $\lambda_m, m\in \mathcal{M}$ to be uniform, and has no additional hyper-parameters. We compare CADP with two prior MMDP algorithms: WSU, MVP~\cite{steimle2021multi} described in \cref{sec:framework}. We also compare CADP with two new gradient-based MMDP methods: mirror descent and natural gradient descent~\cite{bhandari2021linear}, which use the gradient derived in \cref{thm:policy gradient}.

We also compare CADP with applicable algorithms designed for models other than MMDPs. A natural algorithm for solving MMDPs is to reduce them to POMDPs. Therefore, we compare CADP with QMDP approximate solver~\cite{littman1995learning} and BasicPOMCP solver~\cite{silver2010monte} for POMDP planning. Recall that POMDP algorithms compute history-dependent policies, which are more complex but could, in principle, outperform Markov policies. Another method for solving MMDPs is to treat them as Bayesian exploration problems. We, therefore, also compare CADP with MixTS~\cite{hong2022thompson}, which uses Thompson sampling to compute history-dependent randomized policies. The original MixTS algorithm assumes that one does \emph{not} observe the current state and only observes the rewards; we adapt it to our setting in the appendix. All algorithms were implemented in Julia 1.7, and the source code is available at \href{https://github.com/suxh2019/CADP}{https://github.com/suxh2019/CADP}.

\begin{figure}
  \centering
\includegraphics[width=0.8\linewidth]{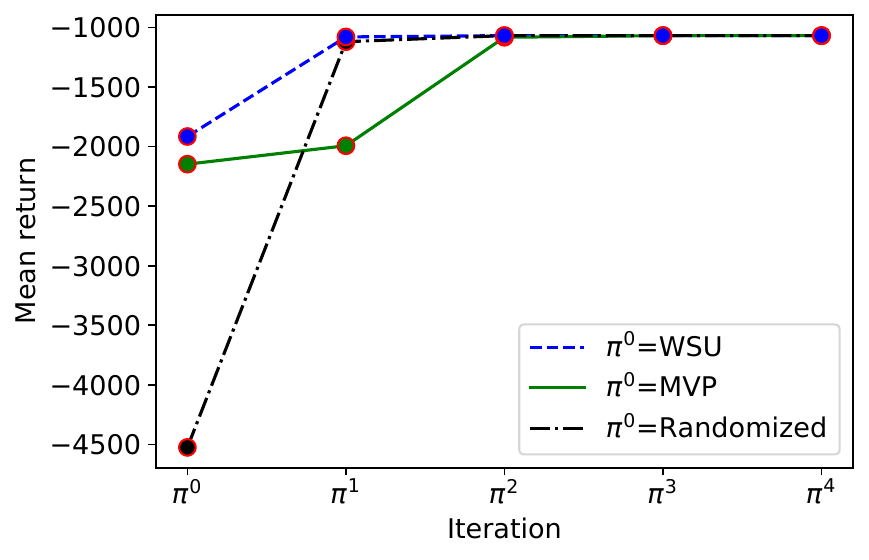}
\caption{Mean Returns of CADP with Different Initial Policies at Different Iterations}
\label{fig:returns}
\end{figure}

\paragraph{Return} First, \cref{fig:returns} compares the mean returns attained in the CADP computation, initialized with WSU, MVP and a randomized policy change with different iterations on domain $POPS$. From the 3rd iteration on, the mean returns of CADP with the three initial policies are essentially identical. The only difference is that CADP initialized with WSU terminates one iteration earlier.  

Second, \cref{tab:returns} summarizes the returns, or solution quality, of the algorithms on
HIV domain for horizon length $T=15$ and other domains for horizon length $T = 50$ evaluated on the \emph{test} set. \Cref{standard-deviation} summarizes the standard deviation of returns of the algorithms on five domains. The results for other time horizons, which are reported in the appendix, are very similar. The algorithm ``Oracle'' describes an algorithm that knows that the true model and its returns are the means of each model's optimal values. Note that Oracle's return may be a lose upper bound on the best possible return. If the runtime of a method is greater than 900 minutes, the method is considered to fail to find a solution and is marked with `--'. BasicPOMCP and QMDP approximate solvers fail to find solutions on domains POP, POPS, and INV.

\begin{table} 
\centering
 \caption{Mean return $\rho(\pi)$ on the Test Set of Policies $\pi$ Computed by Each Algorithm. HIV Values are in 1000s.} 
\begin{tabular}{lrrrrr}
\toprule
  \textbf{Algorithm}  & \textbf{RS} & \textbf{POP} & \textbf{POPS} & \textbf{INV} & \textbf{HIV}\\
  \midrule
  \textbf{CADP}    & \textbf{204}   &\textbf{-361} &\textbf{-1067} &323  &\textbf{42}\\
  WSU              &203    &-542   &-1915 & 323  &42 \\ 
  MVP              & 201   &-704   &-2147 & 323  &42 \\ 
 \midrule
  Mirror           & 181    &-1650  &-3676 &314  &\textbf{42} \\
  Gradient         & 203    &-542   &-1915 &323  &42 \\
 \midrule
  MixTS            & 167    &-1761 &-2857 & \textbf{327} & -1 \\
  QMDP             & 190    & --     & --     &  --   &40 \\
  POMCP            & 58     & --     & --     & --    &30 \\
  \midrule
  Oracle           & 210   &-168  &-882  &332  &53\\
  \bottomrule
\end{tabular}
\label{tab:returns}
\end{table}

\begin{table}
\centering
\caption{Standard deviation of returns of algorithms. HIV values are in 1000s.}
\label{standard-deviation}
\begin{tabular}{lrrrrr}
  \toprule
  \textbf{Algorithm}  & \textbf{RS} & \textbf{POP} & \textbf{POPS} & \textbf{INV} & \textbf{HIV}\\
 \midrule
\textbf{CADP}        &96    &\textbf{1081}  &\textbf{1986}  &\textbf{47}   & \textbf{11} \\ 
WSU                  &98    &1346    &3119  &49   &11   \\ 
MVP                  &89    &2018    &3577  &48   &11    \\ 
\midrule
Mirror              &\textbf{69}    &2150    &4494  &52   &11   \\ 
Gradient            &98    &1346    &3119  &49   &11   \\
 \midrule
MixTS              &223   &4398    &5454  & 58  &26   \\ 
QMDP                &213   &-       &-     &-    &64   \\
 POMCP              &72    &-       & -    & -   &55   \\ 
   \midrule
Oracle      &93   &1032    &1868  &47   &14   \\
  \bottomrule
\end{tabular}
\end{table}

%The results are reported on the test set of models, which is generated from the same posterior distribution over models as the training set.

\paragraph{Run-time}
\Cref{tab:runtimes} summarizes the runtime of several algorithms on the domains for %varying horizon lengths $T$. 
horizon length $T = 50$. All algorithms were executed on a Ubuntu 20.04 with 3.0 GHz Intel processor and 32 GB of RAM. BasicPOMCP and QMDP approximate solvers fail to compute a policy on domains POP, POPS, and INV in a reasonable time. MVP runs fastest. MixTS runs slower than MVP but faster than other algorithms. The CADP method needs several iterations to get the policy that achieves the local maximum. As we expected, the time taken by CADP to solve these instances is several times as much as WSU. How quickly the CADP converges depends on the length of the time horizon and the number of iterations performed.

\begin{table}
    \centering
\caption{Run-times to compute a policy $\pi$ in minutes.}
  \begin{tabular}{lrrrrr}
  \toprule
 \textbf{Algorithm}   & \textbf{RS} & \textbf{POP} & \textbf{POPS} & \textbf{INV}  & \textbf{HIV}\\
  \midrule
 \textbf{CADP}  & 0.29   &69.66  &5.39  & 0.88  & 0.0053\\  
   WSU          & 0.08   & 33.65 &0.95  & 0.45  & 0.0016 \\ 
   MVP          & \textbf{0.05}   &\textbf{27.60}  &\textbf{0.36}  &\textbf{0.22}  &\textbf{0.0002}\\ 
  \midrule
   Mirror       &1.06    & 67.88  & 4.44 & 2.88   & 0.0111 \\ 
   Gradient     & 0.46   & 40.78  & 1.52 & 0.79   & 0.0042\\ 
  \midrule
   MixTS        & 0.07  &28.06   & 0.59  & 0.34   & 0.0018\\ 
   QMDP         & 712   & --     & --    & --     & 0.7100\\ 
   POMCP        & 68    & --     & --    & --     & 0.2066\\
    % \midrule
    % Oracle      & --    & --     & --    & --     & -- \\
  \bottomrule
\end{tabular}
\label{tab:runtimes}
\end{table}
\paragraph{Discussion}
Our results show that CADP consistently achieves the best or near-best return in all domains and time horizons. This is remarkable because it only looks for Markov policies, whereas several other algorithms consider the richer space of history-dependent policies. The computational penalty that CADP incurs compared to just solving the average model, as done by MVP, is only a factor of 3-10. In comparison, state-of-the-art POMDP solvers were unable to solve most of the domains within a factor of 100 of CADP's runtime.

Let us take a closer look at the performance of the algorithms for a horizon $50$ on domain POPS. The runtime of WSU is $0.95$ minutes, but the return that WSU obtains is $-1915$. The runtime taken by CADP is $5.6$ times as much as WSU, but the return for CADP is -1067, which is significantly greater than what WSU achieves. MixTS is a sampling-based algorithm that is guaranteed to perform well over long horizons but has no guarantees for short horizons. However, this assumption may not hold in this domain, which could cause MixTS to perform poorly. The mirror descent algorithm and CADP obtain the same return on the domain HIV, but CADP performs significantly better than the mirror descent algorithm on other domains. Therefore, CADP outperforms those approaches with some runtime penalty. 

\section{Conclusions and Future Work}\label{sec:concl-future-work}
This paper proposes a new efficient algorithm, CADP, which combines a coordinate ascent method and dynamic programming to solve MMDPs. CADP incorporates adjustable weights into the MMDP and adjusts the model weights each iteration to optimize the deterministic Markov policy to the local maximum. Our experiment results and theoretical analysis show that CADP performs better than existing approximation algorithms on several benchmark problems. The only drawback of CADP is that it needs several iterations to obtain a converged policy and increases the computational complexity. In terms of future work, it would be worthwhile to scale up CADP to value function approximation and consider richer soft-robust objectives. It also would be worthwhile to design algorithms that add limited memory to the policy to compute simple history-dependent policies.

\paragraph{Acknowledgements} 
We thank the anonymous reviewers for their comments. This work was supported, in part, by NSF grants 2144601 and 1815275.

% References
\bibliography{su_214}
\end{document}

\onecolumn %% Turn this off if single column is desired for the supplement
\maketitle
% This Supplementary Material should be submitted as a separate file. Please do not append the Supplementary Material to the main paper. 
% Fig. \ref{fig:pitt} and Eq \ref{eq:example} in the main paper can be cross referenced using \texttt{xr}. 
\appendix

\section{Proof of Theorem~\ref{thm:policy gradient}}

\begin{proof}[Proof of \cref{thm:policy gradient}]\label{proof:policy gradient}
  For any time step $\hat{t} \in \mathcal{T}$, we can express the return as 
\begin{align*}
\rho(\pi) &= \mathbb{E}^{\lambda,\pi,p^{\Tilde{m}},\mu} \left[ \sum_{t=1}^{T}  r_{t}^{\Tilde{m}}(\Tilde{s}_{t},\Tilde{a}_{t}) \right] \\
  &= \mathbb{E}^{\lambda,\pi,p^{\Tilde{m}},\mu} \left[ \sum_{t=1}^{\hat{t}-1}  r_{t}^{\Tilde{m}}(\Tilde{s}_{t},\Tilde{a}_{t}) \right]  + \mathbb{E}^{\lambda,\pi,p^{\Tilde{m}},\mu} \left[ \sum_{t=\hat{t}}^{T}  r_{t}^{\Tilde{m}}(\Tilde{s}_{t},\Tilde{a}_{t}) \right] \\
  &\overset{\text{(a)}}{=}  C + \mathbb{E}^{\lambda,\pi,p^{\Tilde{m}},\mu}\left[ \mathbb{E} \left[ \sum_{t=\hat{t}}^{T}  r_{t}^{\Tilde{m}}(\Tilde{s}_{t},\Tilde{a}_{t}) \mid  \tilde{s}_{\hat{t}} , \tilde{m} \right] \right]\\
  &\overset{\text{(b)}}{=}  C + \sum_{m\in \mathcal{M},s_{\hat{t}}\in \mathcal{S},a_{\hat{t}}\in \mathcal{A}}   \mathbb{P}\left[\tilde{m} = m, \tilde{s}_{\hat{t}}= s_{\hat{t}}\right] \pi_{\hat{t}}(s_{\hat{t}}, a_{\hat{t}}) \cdot  \mathbb{E}^{\lambda,\pi,p^{\Tilde{m}},\mu}\left[ \sum_{t=\hat{t}}^{T}  r_{t}^{\Tilde{m}}(\Tilde{s}_{t},\Tilde{a}_{t}) \mid  \tilde{s}_{\hat{t}} = s_{\hat{t}}, \tilde{a}_{\hat{t}} = a_{\hat{t}},  \tilde{m} = m  \right]\\
  &\overset{\text{(c)}}{=}  C + \sum_{m\in \mathcal{M},s_{\hat{t}}\in \mathcal{S},a_{\hat{t}}\in \mathcal{A}}   b_{\hat{t},m}^{\pi}(s_{\hat{t}}) \cdot  \pi_{\hat{t}}(s_{\hat{t}}, a_{\hat{t}}) \cdot  q_{\hat{t},m}^{\pi}(s_{\hat{t}}, a_{\hat{t}})\,.
\end{align*}
Here, we use $C = \mathbb{E}^{\lambda,\pi,p^{\Tilde{m}},\mu} \left[ \sum_{t=1}^{\hat{t}-1}  r_{t}^{\Tilde{m}}(\Tilde{s}_{t},\Tilde{a}_{t}) \right]$ for brevity. The step (a) follows from the law of total expectation, the step (b) follows from the definition of conditional expectation, and the step (c) holds from the definitions of $b$ and $q$ in~\eqref{eq:v-t-m-s},~\eqref{eq:backup}, and~\eqref{eq:belief-state}.

Using the expression above, we can differentiate the return for each $s\in \mathcal{S}$ and $a\in \mathcal{A}$ as
\[
  \frac{\partial \rho(\pi)}{\partial \pi_{\hat{t}}(s,a)}
  \quad =\quad 
  b_{\hat{t},m}^{\pi}(s)  \cdot  q_{\hat{t},m}^{\pi}(s, a)\,,
\]
which uses the fact that $C$, $b_{\hat{t},m}^{\pi}$, and $q_{\hat{t},m}^{\pi}$  are constant with respect to $\pi_{\hat{t}}$. The desired result then holds by substituting $t$ for $\hat{t}$, $\hat{s}$ for $s$, and $\hat{a}$ for $a$.

\end{proof}

\section{Proof of Theorem~\ref{thm:improvement}}

\begin{proof}[Proof of \Cref{thm:improvement}]
  Assume some iteration $n$. The proof then follows directly from the contruction of the policy $\pi^n$ from $\pi^{n-1}$. By the construction in \cref{eq:swsu-optimal-action}, we have that:
  \[
    \rho(\pi_1^{n-1}, \dots\pi_{t-1}^{n-1}, \pi_t^n, \pi_{t+1}^{n}\dots, \pi_T^{n})
    \; \ge\;  
    \rho(\pi_1^{n-1}, \dots\pi_{t-1}^{n-1}, \pi_t^{n-1}, \pi_{t+1}^{n}\dots, \pi_T^{n})\,.
  \]
Note that the optimal form of the policy in \cref{eq:swsu-optimal-action} follows immediately from the standard first-order optimality criteria over a simplex~(e.g,~Ex.~3.1.2 in ~\cite{Bertsekas2016nonlinear}) and the fact that the function optimized in \cref{eq:swsu-optimal-action} is linear (\cref{cor:ret-linear}). In particular, we have that
\[
  \pi_t^n \in \argmax_{\hat{\pi}_t \in \Real^{\mathcal{S} \times\mathcal{A}}}\rho(\pi_1^{n-1}, \dots, \hat{\pi}_t, \dots, \pi_T^{n})
\]
if and only if for each $s\in \mathcal{S}$ and $a\in \mathcal{A}$
\[
  \frac{\partial \rho(\pi_1^{n-1}, \dots, \pi^n_t, \dots, \pi_T^{n})}{\partial\pi_t(s,\pi_t^n(s))}
 \;  \ge  \; 
  \frac{\partial \rho(\pi_1^{n-1}, \dots, \pi^n_t, \dots, \pi_T^{n})}{\partial\pi_t(s,a)}\,.
\]
Intuitively, this means that the optimal policy $\pi^n_t$ must choose actions that have the \emph{maximum} gradient for each state. The optimization in \cref{eq:swsu-optimal-action} then follows by algebraic manipulation from \cref{thm:policy gradient}.

\end{proof}

\section{Proof of Theorem ~\ref{thm:no-sublinear-regret}}

\begin{proof}[Proof of \cref{thm:no-sublinear-regret}]\label{proof:no-sublinear-regret}
Consider the MMDP illustrated in \cref{fig:simple-mmdp-1}. 

\begin{figure}
\centering
\begin{minipage}{.4\textwidth}
\begin{tikzpicture}[->,-Latex,>=Latex,font=\large,node distance=35mm,el/.style = {inner sep=2pt, align=left, sloped},
every label/.append style = {font=\tiny}]
\node[state] (2) {$2$};
\node[state, below of=2] (3) {$3$};
\node[state, below right of=2] (4) {$4$};
\node[state, below left of=2] (1) {$1$};
\draw  (2) edge[left] node[el,above]{$a=1,2$} (1)
  (1) edge[bend right, below] node[el,below,pos=0.7]{$a=1$} (2)
  (1) edge[bend left,below] node[el,above,pos=0.7]{$a=2$} (3)
  (3) edge[bend left,below] node[el,below]{$a=1,2$} (1);
\draw (4) edge[loop,above] node[el,above]{$a=1,2$} (4);
\end{tikzpicture}
\end{minipage}
\begin{minipage}{.4\textwidth}
\begin{tikzpicture}[->,-Latex,>=stealth,font=\large,node distance=35mm,el/.style = {inner sep=2pt, align=left, sloped},
every label/.append style = {font=\tiny}]
\node[state] (2) {$2$};
\node[state, below left of=2] (1) {$1$};
\node[state, below of=2] (4) {$4$};
\node[state, below right of=2] (3) {$3$};
\draw  (2) edge[left]node[el,above]{$a=1,2$} (1)
  (1) edge[bend right, above] node[el,below,pos=0.7]{$a=1$} (2)
  (1) edge[bend left, below] node[el,above,pos=0.7]{$a=2$} (4)
  (3) edge[loop,above] node[el,above]{$a=1,2$} (3)
  (4) edge[bend left, below] node[el,below]{$a=1,2$} (1);
\end{tikzpicture}
\end{minipage}
\caption{Left: model $m_{1}$, right: model $m_{2}$}
\label{fig:simple-mmdp-1}
\end{figure}

First, we describe the time steps, states, rewards, and actions for this MMDP. This MMDP has three time steps, four states $\mathcal{S} = \{1,2,3,4 \}$, two actions $\mathcal{A} = \{1,2\}$, and two models $\mathcal{M} = \{1,2\}$. The model weight for $m_{1}$ is $\lambda$, then the model weight for $m_{2}$ is $1-\lambda$.  State $1$ is the only initial state. In model $m_{1}$, the only non-zero reward 2 is received upon reaching state $2$. The agent takes action $1$, which leads to a transition to state $2$ with a probability of 1. The agent takes action $2$, which leads to a transition to state $3$ with probability 1.  The agent takes action $1$ or $2$ in state $2$, which leads to a transition to state $1$ with probability 1. The agent takes action $1$ or $2$ in state $3$, which leads to a transition to state $1$ with probability 1. The agent takes action $1$ or $2$ in state $4$, which leads to a transition to state $4$ with probability 1. 

In model $m_{2}$, the agent receives rewards 3 upon reaching state $4$ and receives rewards 2 upon reaching state $2$. The agent takes action $1$, which leads to a transition to state $2$ with probability 1. The agent takes action $2$, which leads to a transition to state $4$ with probability 1. The agent takes action $1$ or $2$ in state $2$, which leads to a transition to state $1$ with probability 1.  The agent takes action $1$ or $2$ in state $4$, which leads to a transition to state $1$ with probability 1.  The agent takes action $1$ or $2$ in state $3$, which leads to a transition to state $3$ with probability 1.

Now, let us analyze the regret of this MMDP. The optimal policy of the above example is a history-dependent policy. That is, to take action $2$ at time step $1$. At time step $2$, the agent takes action $1$ or $2$, which leads to a transaction back to state $1$. From time step $3$, if the agent is in model $m_{1}$, then take action $1$; if the agent is in model $m_{2}$, then take action $2$. 

Next, let us analyze the regret of a Markov policy for the MMDP. $S_{t}$ represents a state at time step $t$. State $1$ has two options: select action $1$ or select action $2$. If action $1$ is selected, this will give a regret value of 0 in model $1$ and a regret value of $1$ in model $2$. If action $2$ is selected, this will give a regret value of $2$ in model $1$ and a regret value of $0$ in model $2$. Therefore, at time step $1$, the total regret is 2$\lambda$ or 1$(1-\lambda$). At time step $2$, the agent takes action $1$ or $2$ in state $S_{2}$(1,3 or 4), which leads to a transition back to state $1$, and gets zero rewards and zero regrets. Then repeat the procedure. At time step $3$, the agent can take action $1$ or action $2$ in state $1$ again. For $T=3$, the trajectory of a Markov policy can be $(S_{1} =1, A_{1} = 1, S_{2},A_{2}, S_{3} =1,A_{3} =1)$, $(S_{1} =1, A_{1} = 1, S_{2},A_{2}, S_{3} =1,A_{3} =0)$, $(S_{1} =1, A_{1} = 0, S_{2},A_{2}, S_{3} =1,A_{3} =1)$, or $(S_{1} =1, A_{1} = 0, S_{2},A_{2}, S_{3} =1,A_{3} =0)$. The accumulated regret can be 2$\lambda$ +1$(1-\lambda$), 2$\lambda$ + 2$\lambda$, 1$(1-\lambda$) + 1$(1-\lambda$). That is, the regret is increased by 1$(1-\lambda$) or 2$\lambda$ for every two time steps. 
\[
R_t(\pi) \geq \frac{\min \{2\lambda, 1-\lambda\}}{2} \cdot t 
\]
Let $c = \frac{\min \{2\lambda, 1-\lambda\}}{2} $, $t' \geq 2$, then we always have 
\[
R_t(\pi) \geq c \cdot t \quad \text{for all} \quad  t \geq t'\
\]

No matter which Markovian policy the agent follows, the accumulated regret will be linear with respect to $t$. Therefore, for this MMDP, there exists no Markovian policy that achieves sub-linear regret.
\end{proof}

\section{Adapted MixTS Algorithm}
The adapted MixTS algorithm is formalized in \cref{alg:MixTS}. $P_0$ is the prior of MDPs and follows the uniform distribution. At the beginning of episode $t$, sample a MDP $M_t$ from the posterior $P_t$ and compute a policy $\pi_t$ that maximizes the value of $M_t$. Then at each time step $h$, take the action $A_h$ based on the policy $\pi_t$ and obtain reward $Y_h$. For each MDP $m \in \mathcal{M}$, update its posterior based on the received rewards.
\begin{algorithm}
   \caption{Adapted MixTS}
   \label{alg:MixTS}
   \textbf{Input}: The prior of MDPs $P_0$ 
\begin{algorithmic}[1]
   \STATE Initialize $P_1 \longleftarrow P_0$\\
   \FOR{episodes t =1, $\cdots$ ,$\mathcal{N}$ }
   \STATE Sample $ M_{t}\sim P_{t}$ \\ 
   \STATE Compute $\pi_{t}$ = $\pi^{M_{t}}$ \\
   \FOR{timesteps h = 1, $\cdots$, H }
   \STATE Select $A_{h} \gets \pi_{t}(S_{h})$  \\
   \STATE Observe reward $Y_{h}$
  \STATE Update $P_{t+1}(m) 	\propto P_{t}(m)P(Y_{h} \mid A_{h};m), \forall m \in \mathcal{M}$
   \ENDFOR
    \ENDFOR
  \end{algorithmic}
\end{algorithm}

\section{Numerical Results: Details}

\subsection{Domain Details}
The CSV files of all domains are available at \href{https://github.com/suxh2019/CADP}{https://github.com/suxh2019/CADP}. ``initial.csv''  specifies the initial distribution over states. ``parameters.csv'' contains the discount factor. ``training.csv''and ``test.csv''have the following columns: ``idstatefrom'', ``idaction'', ``idstateto'', ``idoutcome'', ``probability'', and ``reward''. Each row entry specifies a transition from ``idstatefrom'' after taking an action ``idaction" to state ``idstateto'' with the associated ``probability'' and ``reward'' in model ``idoutcome''. A policy is computed from the ``training.csv'', and the policy is evaluated on the ``test.csv''. The models are identified with integer values $0, \cdots, M-1$, and each model is defined on the same state space and the action space. The states are identified with integer values $0, \cdots, S-1$, and the actions are identified with integer values $0, \cdots, A-1$. Note that the number of actions taken in each state $s$ is less or equal to $A$. Each MDP model has its unique reward functions and transition probability functions.

\subsection{Additional Simulation Results}
\Cref{app:returns} shows mean returns of algorithms on five domains at different time steps. \Cref{app:standard-deviation} shows the standard deviations of returns of algorithms on five domains at different time steps. The algorithm “Oracle” knows the true model and its standard deviation summarizes the variability of MDP models in an MMDP. The standard deviations of other algorithms include both the variability of MDP models and the variability of a policy in a MDP model. \cref{app:runtimes} shows runtimes of algorithms on five domains at different time steps. $CADP $ performs best with some runtime penalty.
\begin{table*}
  \centering
  \caption{Mean Returns $\rho(\pi)$ on the Test Set of Policies $\pi$ Computed by Each Algorithm.}\label{app:returns}
\begin{tabular}{lrrrrrrrrrr}
\toprule
  \textbf{Algorithm}  \bfseries  & \multicolumn{2}{c}{\textbf{RS}} \bfseries & \multicolumn{2}{c}{\textbf{POP}}  \bfseries & \multicolumn{2}{c}{\textbf{POPS}}\bfseries & \multicolumn{2}{c}{\textbf{INV}} & \multicolumn{2}{c}{\textbf{HIV}} \\
 \bfseries   &  T = 100  \bfseries & T =150  &  T = 100  \bfseries & T =150  &  T = 100  \bfseries & T =150  &  T = 100  \bfseries & T =150  &  T = 5  \bfseries & T =20\\
  \midrule
 \textbf{CADP}   &\textbf{207}  & \textbf{207}  & \textbf{-368} &\textbf{-368} &\textbf{-1082} &\textbf{-1082} &348 &\textbf{350} &\textbf{33348} &\textbf{42566}\\
   WSU               &206   &206  &-551     &-551   &-1934 &-1932 &347 &349   &\textbf{33348} &42564\\
 MVP               &204   &204  &-717     &-717   &-2178 &-2179 &348 &\textbf{350}  &\textbf{33348} &42564 \\ 
\midrule
 Mirror            &183   &183  &-1601    &-1600  &-3810 &-3800 &343 &345   &\textbf{33348} &\textbf{42566}\\
 Gradient          &206   &206  &-551     &-551   &-1934 &-1932 &347 &349   &\textbf{33348} &42564\\
\midrule
 MixTS             &172   &176  &-1961    &-1711  &-3042 &-3016 &\textbf{350} &\textbf{350}  &293 &-1026\\
QMDP               &201   &183  & -       &-      &-     &-     &-   &-     &30705  &39626\\ 
POMCP              &54      & 64    & -       &-      &-     &-     &-   &-     &25794  &30910\\
  \midrule
 Oracle     &213  & 213 &-172     &-172   &-894  &-894  &358 &360   &40159  &53856\\
  \bottomrule
\end{tabular}
\end{table*}

\begin{table*}
  \centering
  \caption{Standard Deviation of Returns of Algorithms on Five Domains.}\label{app:standard-deviation}
\begin{tabular}{lrrrrrrrrrr}
\toprule
  \textbf{Algorithm}  \bfseries  & \multicolumn{2}{c}{\textbf{RS}} \bfseries & \multicolumn{2}{c}{\textbf{POP}}  \bfseries & \multicolumn{2}{c}{\textbf{POPS}}\bfseries & \multicolumn{2}{c}{\textbf{INV}} & \multicolumn{2}{c}{\textbf{HIV}} \\
 \bfseries   &  T = 100  \bfseries & T =150  &  T = 100  \bfseries & T =150  &  T = 100  \bfseries & T =150  &  T = 100  \bfseries & T =150  &  T = 5  \bfseries & T =20\\
\midrule
\textbf{CADP}     &98   &98   &\textbf{1095}   & \textbf{1095}     &\textbf{2007} &\textbf{2007}  &\textbf{51}  &\textbf{51} & 9342 &\textbf{11309} \\
 MVP                &90   &90   &2046   &2046  &3619   &3620  &52  &52 &\textbf{7729} &12234 \\
 WSU                &100  &100  &1364   &1364  &3147   &3146  &53  &53 &\textbf{7729}  & 12234\\
\midrule
 Mirror             &\textbf{70}  &\textbf{70}   &2081   &2081  &4534   &4530  &57  &58 &\textbf{7729} &12237 \\
 Gradient           &100  &100  &1364   &1364  &3147   &3146  &53  &53 &\textbf{7729}  & 12234\\
  \midrule
MixTS             &226  &231   &4436  &4187   &5507   &5542  &58  &58 &23689  &27792 \\
QMDP               &193  &204   & -     &-     &-      &-     &-   &-  &42987  &61596 \\
POMCP              &66   &118   &-      &-     &-      &-     &-   &-  &42208  &57772 \\
  \midrule
 Oracle    &95   &95   &1045   &1045  &1889   &1889  &51  &51 &9029  &14796 \\
  \bottomrule
\end{tabular}
\end{table*}

\begin{table*}
  \centering
  \caption{Run-times of Algorithms on Five Domains in Minutes.}\label{app:runtimes}
\begin{tabular}{lrrrrrrrrrr}
\toprule
  \textbf{Algorithm}  \bfseries  & \multicolumn{2}{c}{\textbf{RS}} \bfseries & \multicolumn{2}{c}{\textbf{POP}}  \bfseries & \multicolumn{2}{c}{\textbf{POPS}}\bfseries & \multicolumn{2}{c}{\textbf{INV}} & \multicolumn{2}{c}{\textbf{HIV}} \\
 \bfseries   &  T = 100  \bfseries & T =150  &  T = 100  \bfseries & T =150  &  T = 100  \bfseries & T =150  &  T = 100  \bfseries & T =150  &  T = 50  \bfseries & T =100\\
\midrule
 MVP               &\textbf{0.05}  &\textbf{0.05}  &\textbf{27.68}   &\textbf{27.51}   &\textbf{0.36}   &\textbf{0.36}   &\textbf{0.22}  &\textbf{0.22}   &\textbf{0.0003}  &\textbf{0.0003}\\
 WSU               &0.12  &0.14  &40.02   &45.39   &1.53   &2.37   &0.67  &0.89   &0.0033  &0.0048\\
\textbf{CADP}              &0.52  &1.13  &124.39  &173.04  &12.12  &16.21  &1.53  &2.22   &0.0109  &0.0164\\
\midrule
 Mirror            &1.86  &3.11  &113.08  &158.06  &8.08   &11.90  &35.90 &53.6   &0.0221  &0.0330\\
 Gradient          &0.51  &0.74  &56.82   &69.32   &2.97   &4.31   &1.12  &1.44   &0.0083  &0.0123\\
  \midrule
MixTS             &0.09  &0.12  &32.08   &35.36   &0.80   &1.03   &0.47  &0.59   &0.0033  &0.0047\\
QMDP               &712   & 712  & -      &-       &-      &-      &-     & -     &0.7071  &0.7071\\
POMCP             &68    &68    & -      &-       &-      &-      &-     &-      &0.2066  &0.2066\\
  \bottomrule
\end{tabular}
\end{table*}